\documentclass{article}
\usepackage{ijcai19}

\usepackage{times}
\usepackage{soul}
\usepackage{url}
\usepackage[hidelinks]{hyperref}
\usepackage[utf8]{inputenc}
\usepackage{graphicx}
\usepackage{amsmath}
\usepackage{booktabs}
\usepackage{algorithm}
\usepackage{algorithmic}
\usepackage{bm}
\usepackage{amsfonts}
\usepackage{multirow}
\usepackage{amssymb}
\usepackage{subfigure}
\usepackage{wrapfig}

\DeclareMathOperator*{\argmax}{arg\,max}

\newcommand{\algo}{{\sc\textsf{FailMaker-AdvRL}}}

\usepackage[font=small]{caption}

\title{Failure-Scenario Maker for Rule-Based Agent using Multi-agent Adversarial \\Reinforcement Learning and its Application to Autonomous Driving}

\author{
    Akifumi Wachi
    \affiliations
    IBM Research AI \emails
    akifumi.wachi@ibm.com
}

\begin{document}

\maketitle

{\let\thefootnote\relax\footnote{{Supplemental material: \url{http://bit.ly/2IpcCQN}}}}

\begin{abstract}
We examine the problem of adversarial reinforcement learning for multi-agent domains including a rule-based agent. Rule-based algorithms are required in safety-critical applications for them to work properly in a wide range of situations. Hence, every effort is made to find failure scenarios during the development phase. However, as the software becomes complicated, finding failure cases becomes difficult. Especially in multi-agent domains, such as autonomous driving environments, it is much harder to find useful failure scenarios that help us improve the algorithm. We propose a method for efficiently finding failure scenarios; this method trains the adversarial agents using multi-agent reinforcement learning such that the tested rule-based agent fails. We demonstrate the effectiveness of our proposed method using a simple environment and autonomous driving simulator.
\end{abstract}

\section{Introduction}
If the decision-making algorithm in safety-critical applications does not work properly, the resulting failure may be catastrophic. To prevent such results occurring after deployment, we must determine as many failure cases as possible and then improve the algorithm in the development phase \cite{doi:10.1080/001401398186568}. Autonomous driving and flight algorithms especially must work properly in a multi-agent environment \cite{urmson2008autonomous,kim2003flight}, which requires us to craft adversarial situations for the tested algorithm by incorporating the interactions with other agents.

Reinforcement learning (RL) has recently achieved significant results; examples range from robotics manipulation \cite{levine2016end} to game playing \cite{mnih2015human,silver2016mastering}. However, most of the software in the practical applications are still rule-based because of the explainability or backwards compatibility. This is true with autonomous driving algorithms as well; hence, we need the algorithms to craft adversarial situations for the rule-based algorithm.

\begin{figure}[t]
    \centering
    \includegraphics[width=65mm]{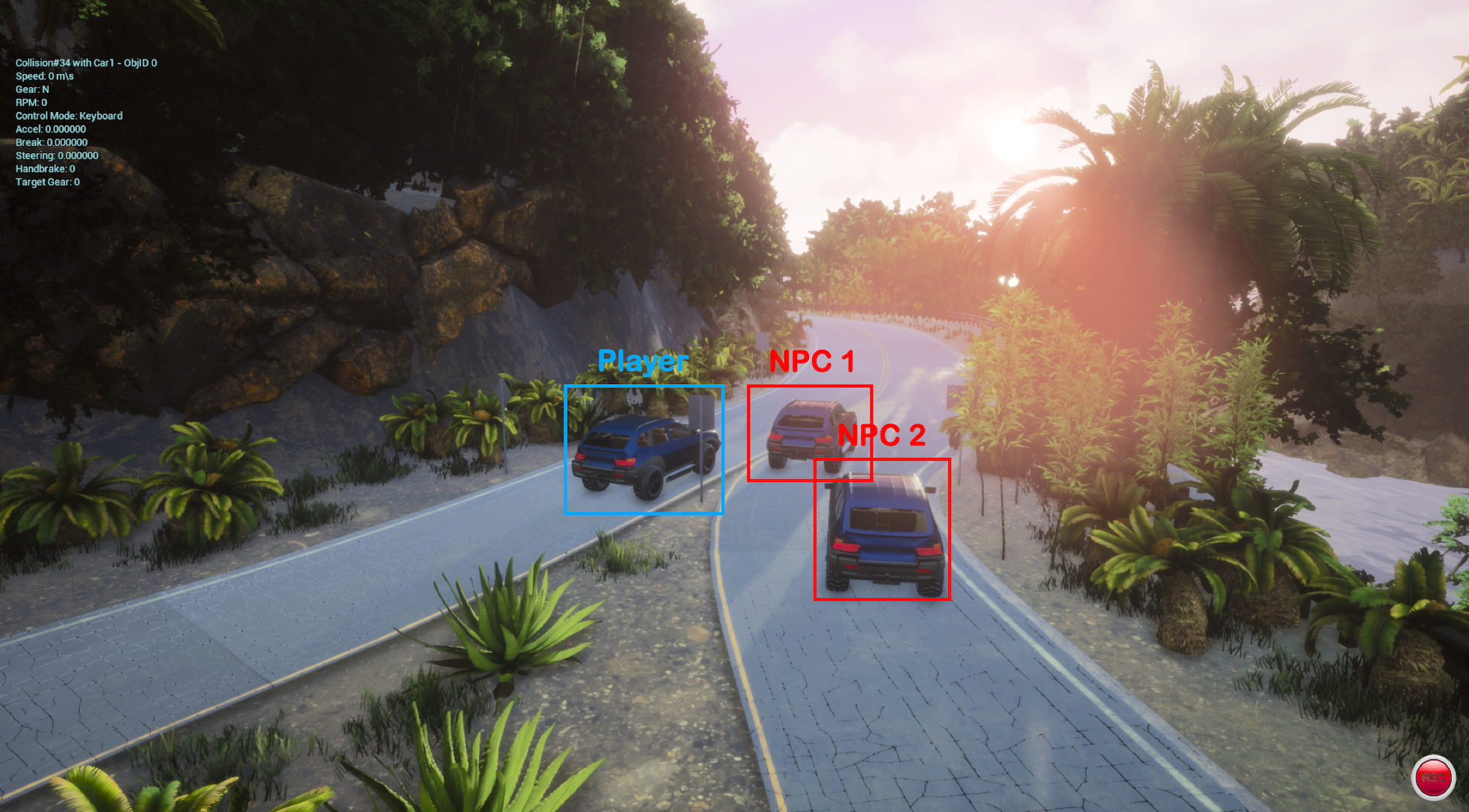}
    \caption{Conceptual image. A blue car is a rule-based player, and two red cars are adversarial RL-based NPCs.}
    \label{fig:marl_airsim}
\end{figure}

As such, it makes sense to train adversarial RL-based agents (i.e., \textit{non-player characters, NPCs}) such that the agent with the tested rule-based algorithm (i.e., \textit{player}) fails. 
By training NPCs in RL frameworks, we create various adversarial situations without specifying the details of the NPCs' behaviors.
We focus on the decision-making aspect rather than image recognition; hence, the failure means collisions in most cases.
Figure~\ref{fig:marl_airsim} gives a conceptual image of the agents in our research. By training NPCs (red) in an adversarial manner, we aim to obtain the failure cases of the player (blue).

In this problem set, however, we encounter the following four problems. 
First, pure adversarial training often results in obvious and trivial failure. For example, if multiple NPCs intentionally try to collide with the player, the player will surely fail; however, such failure cases are useless for improving the rule-based algorithm. 
Second, when the player fails, it is not always clear which NPCs induce the failure. For efficient and stable learning, we should present the adversarial reward only for the NPCs that contribute to the player's failure.
Third, the player does not have a notion of reward, and it is often possible to know only whether or not the player fails. That is, from the perspective of the NPCs, the reward for the player's failure is extremely sparse (e.g., if the player fails, NPCs get the adversarial reward of ``1''; otherwise, they get ``0'').\footnote{We have an option to define a \textit{virtual} reward for the player. However, it is often difficult to precisely define the (virtual) reward.} 
Finally, if the player rarely fails, NPCs are trained using the imbalanced past experience. The imbalanced experience, dominated by the player's success scenarios, prevents NPCs from acquiring a good policy.

\subsubsection*{Contributions}
We propose a novel algorithm, \algo; this approach trains the adversarial RL-based NPCs such that the tested rule-based player fails. In addition, to address the four problems discussed above, we present the following ideas. 

\vspace{2pt}
\noindent
\textbf{Adversarial learning with personal reward.} To train NPCs to behave in an adversarial but natural manner, we consider personal reward for NPCs as well as the adversarial reward. This is because we consider that, if it behaves unnaturally, an NPC itself loses the personal reward. When an NPC tries to collide with the player, the personal reward is lost. NPCs should have their own objective (e.g., \textit{safely} arrive at the goal as early as possible), and we consider the loss of the personal reward to ensure natural behavior.

\vspace{2pt}
\noindent
\textbf{Contributor identification (CI) and adversarial reward allocation (AdvRA).} To identify the NPCs that should be provided with an adversarial reward, we propose contributor identification (CI). This algorithm classifies all the NPCs into several categories depending on their degree of contribution by re-running the simulation with the subsets of NPCs. In addition, to handle sparse (episodic) adversarial reward, we propose adversarial reward allocation (AdvRA), which properly allocates the adversarial reward to the NPCs. This algorithm allocates the sparse adversarial reward among each state and action pair that contributes to the player's failures. 

\vspace{2pt}
\noindent
\textbf{Prioritized sampling by replay buffer partition (PS-RBP).} To address the problem caused by imbalanced experience, we partition the replay buffer depending on whether or not the player succeeds, and then train the NPCs using the experience that is independently sampled from the two replay buffers.

\vspace{2pt}
We demonstrated the effectiveness of \algo\ with two experiments using both a simple environment and a 3D autonomous driving simulator.

\section{Related Work} 

This work is on adversarial multi-agent RL. We review previous work on \textit{adversarial learning} and \textit{multi-agent RL}.

\paragraph{Adversarial learning.} Adversarial learning can be frequently seen in computer vision tasks. For example, \cite{szegedy2013intriguing} and \cite{kurakin2016adversarial} aimed to craft adversarial images that make an image classifier work improperly. 
In addition, many previous works constructed a more robust classifier by leveraging generated adversarial examples, generative adversarial net (GAN) \cite{goodfellow2014generative} being a representative example of such work. 

\paragraph{Multi-agent reinforcement learning (MARL).} The relationship among multiple agents can be categorized into cooperative, competitive, and both. Most previous work on MARL addresses cooperative tasks, in which the cumulative reward is maximized as a group \cite{lauer2000algorithm,panait2005cooperative}. In particular, \cite{foerster2017counterfactual} proposed a method with a centralized critic for a fully cooperative multi-agent task.
Algorithms on MARL applicable with competitive settings have recently been proposed by \cite{pinto2017robust,lowe2017multi,li2019robust}. \cite{pinto2017robust} consider a two-player zero-sum game in which the protagonist gets a reward $r$ while the adversary gets a reward $-r$. In \cite{lowe2017multi}, a centralized critic approach called multi-agent deep deterministic policy gradient (MADDPG) is proposed for mixed cooperative and competitive environments; MADDPG is a similar idea as that in \cite{foerster2017counterfactual}. 

\paragraph{Other related work.} Our proposed AdvRA and CI are varieties of credit assignment methods \cite{devlin2014potential,nguyen2018credit}. Most existing methods allocate the proper reward to each agent by utilizing difference reward \cite{wolpert2002optimal} under the assumption that all agents are trained by RL. Also, prioritized experience-sampling techniques were previously proposed in \cite{schaul2015prioritized} and \cite{horgan2018distributed}; these techniques enable efficient learning by replaying important experience frequently.
For previous work on autonomous driving testing, we refer readers to \cite{pei2017deepxplore} and \cite{tian2018deeptest}. Many previous studies have addressed test-case-generation problems related to image recognition rather than the decision making, which we address in this work.

\section{Background}
\label{sec:background}

In this section, we briefly review the standard Markov games and previous studies that underline \algo. The definitions and notations follow \cite{lowe2017multi}.

\subsection{Markov Games}
Partially observable Markov games are a multi-agent extension of Markov decision processes \cite{littman1994markov}.
A Markov game for $N$ agents is defined by a set of states $\mathcal{S}$ meaning all the possible configuration of all agents, a set of actions $\mathcal{A}_1, \ldots, \mathcal{A}_N$, and a set of observations $\mathcal{O}_1, \ldots, \mathcal{O}_N$. Agent $i$ decides the action using a stochastic policy $\bm{\pi}_{\theta_i}: \mathcal{O}_i \times \mathcal{A}_i \rightarrow [0, 1]$, which produces the next state depending on the state transition function $\mathcal{T}: \mathcal{S} \times \mathcal{A}_1 \times \ldots \times \mathcal{A}_N \rightarrow \mathcal{S}$. Each agent $i$ obtains the reward of $r_i: \mathcal{S} \times \mathcal{A}_i \rightarrow \mathbb{R}$ in accordance with the state and agents' action and acquires a new observation $\bm{o}_i: \mathcal{S} \rightarrow \mathcal{O}_i$. The initial states are determined by a probabilistic distribution $\rho: \mathcal{S} \rightarrow [0,1]$. Each agent $i$ tries to maximize the expected cumulative reward, $R_i = \sum_{t=0}^T\gamma^t r_i^t$, where $\gamma$ is a discount factor, and $T$ is the time horizon.

\subsection{Policy Gradient (PG) and Deterministic Policy Gradient (DPG)}
Policy gradient (PG) algorithms are popular in RL tasks. The key idea of PG is to directly optimize the parameters $\theta$ of policy $\pi$ to maximize the expected cumulative reward by calculating the policy's gradient with regard to $\theta$. 

Deterministic policy gradient (DPG) algorithms are variants of PG algorithms that extend the PG framework to deterministic policy, $\bm{\mu}_\theta: \mathcal{S} \rightarrow \mathcal{A}$. In the DPG framework, the gradient of the objective function $J(\theta) = \mathbb{E}_{s \sim \rho^{\bm{\mu}}}[R(s,a)]$ is written as:
\[
\nabla_\theta J(\theta) = \mathbb{E}_{s \sim \mathcal{D}}[\nabla_\theta \bm{\mu}_\theta (s) \nabla_a Q^{\bm{\mu}} (s,a)|_{a=\bm{\mu}_\theta(s)}],
\]
where $\mathcal{D}$ is the replay buffer that contains the tuple, $(\bm{s}, a, r, \bm{s}')$.
Deep deterministic policy gradient (DDPG) approximates the deterministic policy $\bm{\mu}$ and critic $Q^{\bm{\mu}}$ using deep neural networks.

\subsection{Multi-Agent Deep Deterministic Policy Gradient (MADDPG)}

MADDPG is a variant of DDPG for multi-agent settings in which agents learn a centralized critic on the basis of the observations and actions of all agents \cite{lowe2017multi}. 

More specifically, consider a game with $N$ agents with policies parameterized by $\bm{\theta} = \{\theta_1, \ldots, \theta_N \}$, and let $\bm{\mu} = \{ \bm{\mu}_1, \ldots, \bm{\mu}_N \} = \{ \bm{\mu}_{\theta_1}, \ldots, \bm{\mu}_{\theta_N} \}$ be the set of all agents’ policies. Then the gradient of the expected return for agent $i \in [1,N]$ with policy $\bm{\mu}_i$, $J(\theta_i) = \mathbb{E}[R_i]$ is written as:
\begin{alignat*}{2}
\nabla_{\theta_i} J(\theta_i) = \mathbb{E}_{\bm{x}, a \sim D} [ \nabla_{\theta_i} \bm{\mu}_i(o_i) \nabla_{a_i} Q_i^{\bm{\mu}}(\bm{x}, \bm{a}_{1:N}) |_{a_i=\bm{\mu}_i(o_i)}],
\end{alignat*}
where $\bm{a}_{1:N} = \{a_1, \ldots, a_N\}$. For example, $\bm{x}$ is defined as $\bm{x} = (o_1, \ldots, o_N)$. The replay buffer $\mathcal{D}$ contains the tuples $(\bm{x}, \bm{x}', a_{1:N}, r_1, \ldots, r_N )$, recording the experiences of all agents. 
The action-value function $Q_i^{\bm{\mu}}$ is updated as:
\begin{alignat*}{2}
\mathcal{L}(\theta_i) &= \mathbb{E}_{\bm{x}, a, r, \bm{x}'} [(Q_i^{\bm{\mu}}(\bm{x}, \bm{a}_{1:N})-y)^2], \\
y &= r_i + \gamma Q_i^{\bm{\mu}'}(\bm{x}', \bm{a}'_{1:N})|_{a'_l=\bm{\mu}'_l(o_l)},
\end{alignat*}
where $\bm{\mu}' = \{\bm{\mu}_{\theta'_1}, \ldots, \bm{\mu}_{\theta'_N} \}$ is the set of target policies with
delayed parameters $\theta'_i$.

\section{Problem Statement}
\label{sec:problem_statement}

We consider a multi-agent problem with a rule-based player and single or multiple NPCs.\footnote{Some of the key ideas in this paper can be employed in problem settings where the player is an RL agent.} At every time step, the player chooses an action on the basis of the deterministic rule (i.e. tested algorithm). Our objective is to train the NPCs adversarially to create situations in which the player makes a mistake. 

We model this problem as a subspecies of multi-agent Markov games. For a player and $N$ NPCs, this game is defined by a set of states $\mathcal{S}$ meaning all the possible configuration of agents, a set of actions $\mathcal{A}_0, \mathcal{A}_1, \ldots, \mathcal{A}_N$, and a set of observations $\mathcal{O}_0, \mathcal{O}_1, \ldots, \mathcal{O}_N$. In the rest of this paper, the subscript ``0'' represents variables that are for the player. The player chooses an action on the basis of the deterministic rule $\bm{\mu}_0$, which does not evolve throughout the training of NPCs.\footnote{We assume that $\bm{\mu}_0$ consists of the numerous number of such deterministic rules as ``if the signal is red, then stop.''} Each NPC uses a stochastic policy $\bm{\pi}_{\theta_i}: \mathcal{O}_i \times \mathcal{A}_i \rightarrow [0, 1]$. The next state is produced depending on the state transition function $\mathcal{T}: \mathcal{S} \times \mathcal{A}_0 \times \mathcal{A}_1 \times \ldots \times \mathcal{A}_N \rightarrow \mathcal{S}$. The player does not obtain a reward from the environment but receives a new observation $\bm{o}_0: \mathcal{S} \rightarrow \mathcal{O}_0$. Each NPC $i$ obtains a personal reward as the function $\hat{r}_i: \mathcal{S} \times \mathcal{A}_i \rightarrow \mathbb{R}$ in accordance with the state and NPCs' action, and at the same time, receives a new observation $\bm{o}_i: \mathcal{S} \rightarrow \mathcal{O}_i$. The initial states are determined by a probabilistic distribution $\rho: \mathcal{S} \rightarrow [0,1]$. 
The ultimate goal of the NPCs is to make the player fail. At the end of each episode, NPCs obtain the binary information on whether or not the player failed.

\section{Method}

\begin{algorithm}[t]
\caption{\textbf{\space \algo}}
\label{algorithm1}
\begin{algorithmic}[1]
\FOR {$e = 1$ to $M$}
\STATE Initialize a random process $\mathcal{N}$ for action exploration; then, obtain the initial state $\bm{x}$
\FOR {$t = 1$ to $T$}
\STATE The player executes the action $a_0$ on the basis of the rule $\bm{\mu}_0$.
\STATE For the adversarial NPC $i$, select the action $a_i = \bm{\mu}_{\theta_i}(o_i) + \mathcal{N}_t$
\STATE NPCs execute the actions $a=(a_1, \ldots, a_N)$ and observe their personal reward $\hat{r}$ and new state $\bm{x}'$
\STATE Store $(\bm{x}, a, \hat{r}, \bm{x}')$ in temporal replay buffer $\mathcal{D}_\text{tmp}$
\STATE $\bm{x} \leftarrow \bm{x}'$
\ENDFOR
\IF {the player succeeds}
\STATE Samples in $\mathcal{D}_\text{tmp}$ are moved to $\mathcal{D}^\text{+}$
\STATE $r_i^t \leftarrow \hat{r}_i^t$
\ELSE
\STATE After executing CI and AdvRA, the samples in $\mathcal{D}_\text{tmp}$ are moved to $\mathcal{D}^-$
\STATE $r_i^t \leftarrow \hat{r}_i^t + \alpha \cdot \bar{r}_i^t$ 
\ENDIF
\FOR {NPC $i=1$ to $N$}
\STATE Randomly choose $\eta(e) S$ samples from $\mathcal{D}^\text{+}$ and $(1-\eta(e))S$ samples from $\mathcal{D}^-$ and create a random minibatch of $S$ samples with PS-RBP
\STATE Update the critic and actor
\ENDFOR
\STATE Update the target network parameters for each NPC $i$
\ENDFOR
\end{algorithmic}
\end{algorithm}

Algorithm~\ref{algorithm1} outlines \algo. At each iteration, the agents execute the actions on the basis of their policy (Line $2-9$). If the player succeeds in the episode, the experience is stored in $\mathcal{D}^+$. If not, the experience is stored in $\mathcal{D}^-$ after identifying the contributing NPCs and allocating the appropriate adversarial reward (Line $10-16$). Finally, the NPCs are adversarially trained using the experiences that are independently sampled from $\mathcal{D}^+$ and $\mathcal{D}^-$ (Line $17-21$). In this section, we first explain three key ideas and then describe the overall algorithm. The detailed pseudocode is given in the supplemental material.

\subsection{Adversarial Learning with Personal Reward}
When we simply train NPCs in an adversarial manner, they will try to make the player fail in whatever way they can. This often results in unnatural situations in which all NPCs try to collide with the player. Considering real applications, it is essentially useless to obtain such unnatural failure cases. 

We obtain the player's failure cases by considering a personal reward for NPCs as well as the adversarial reward. For, we consider unnatural situations to be ones in which NPCs themselves lose a large personal reward. Therefore, we train the NPCs while incentivizing them to maximize the cumulative personal and adversarial reward. Let $\hat{r}_i^t$ and $\bar{r}_i^t$ denote the NPC $i$'s personal and adversarial reward. The reward is written as
\begin{equation}
\label{eq:reward_consistency}
r_i^t = \hat{r}_i^t + \alpha \cdot \bar{r}_i^t,
\end{equation}
where $\alpha \in \mathbb{R}$ is the scaling factor. In Section~\ref{sec_advra}, we will explain how to design the adversarial reward, $\bar{r}_i^t$.

\subsection{Contributors Identification (CI) and Adversarial Reward Allocation (AdvRA)}
\label{sec_advra}
When the player fails in an episode, the NPCs receive the adversarial reward. How, though, should we reward each NPC?  For efficient and stable training, we should reward the state and action pairs that contribute to the player's failure.

\paragraph{Contributors identification (CI).}
First, to restrict the NPCs that obtain the adversarial reward, we identify the NPCs that contributed to the player's failure. 
More precisely, for $K \in \mathbb{N}$, we classify NPCs into class $k$ contributors ($k=1, 2, \ldots, K$). 
Specifically, class 1 contributors can foil the player alone, and class 2 contributors can foil the player with another class 2 contributor (though they are not class 1 contributors). 
To identify the class 1 contributors, we re-run the simulation with the player and a single NPC. If the player fails, the NPC is classified as a class 1 contributor. 
Next, to identify class 2 contributors within the NPCs excluding for the class 1 contributors, we re-run the simulation with the player and two NPCs while seeding the randomness of the simulation.  
The above steps are continued until we identify the class $K$ contributors.
In the rest of the paper, we denote $\mathcal{C}_k$ as the set of class $k$ contributors and $\mathcal{C}$ as the set of contributors; that is, $\mathcal{C}=\mathcal{C}_1 \cup \ldots \cup \mathcal{C}_K$. When identifying the class $k$ contributors, the number of simulations is ${}_{N} C_k$; hence, $K$ should be small for a large $N$. Practically, since traffic accidents are caused by the interaction among a small number of cars, setting small $K$ does not usually affect the practicality.

\paragraph{Adversarial reward allocation (AdvRA).}
After identifying the contributors, we assign the adversarial reward to each state and action pair. Let $g$ denote the measure of the contribution to the player's failure, which we will call the \textit{contribution function}. Suppose NPC $i^*$ is the biggest contributor to the player's failure. The biggest contributor is identified by:
\[
i^* = \argmax_{i \in \mathcal{C}_{k_\text{min}}} \left(\max_t g_t^i(\bm{x}, a_i)\right),
\]
where $k_\text{min} = \min \{k \in [1,K] \mid \mathcal{C}_k \ne \emptyset \}$.
Here, we allocate the adversarial reward to each state and action pair of the NPCs as follows. First, the adversarial reward of ``1'' is allocated to each state and action pair of NPC $i^*$ depending on the contribution function $g$:
\begin{equation*}
    \bar{r}_{i^*}^t(\bm{x}, a_{i^*}) = g_{i^*}^t(\bm{x}, a_{i^*})/\sum_t g_{i^*}^t(\bm{x}, a_{i^*}).
\end{equation*}
The adversarial reward is then allocated to the all contributors, $\mathcal{C}$ in accordance with their contributions. For NPC $i \in \mathcal{C}_k$, we set
\begin{equation}
\bar{r}_i^t(\bm{x}, a_i) = w(k) \cdot \bar{r}_{i^*}^t(\bm{x}, a_{i^*}) \cdot g_i^t(\bm{x}, a_i)/g_{i^*}^t(\bm{x}, a_{i^*}),
\label{eq:advra_2}
\end{equation}
where $w(k) \in [0, 1]$ is the monotone non-increasing function with regard to $k$, which represents the weight between the classes of the contributors. Empirically, to prevent NPCs from obstructing other NPCs, $w$ should not be too small. 
In our simulation, the contribution function $g_i^t$ is simply defined using the distance between the player and NPC $i$; that is,
\begin{equation}
\label{eq:contribution_func}
g_i^t = \exp(-\beta \cdot \|\bm{s}_0^t - \bm{s}_i^t\|),
\end{equation}
where $\beta$ is a positive scalar, and $\bm{s}_0^t$ and $\bm{s}_i^t$ are the positions of the player and NPC $i$, respectively. Intuitively, this contribution function is based on the consideration that the action inducing the player's failure should be taken when NPC $i$ is close to the player. An alternative would be to define the contributor function using commonly used safety metrics such as headway and time-to-collision \cite{vogel2003comparison}.

\subsection{Prioritized Sampling by Replay Buffer Partition (PS-RBP)}

\begin{figure}
    \centering
    \includegraphics[width=80mm]{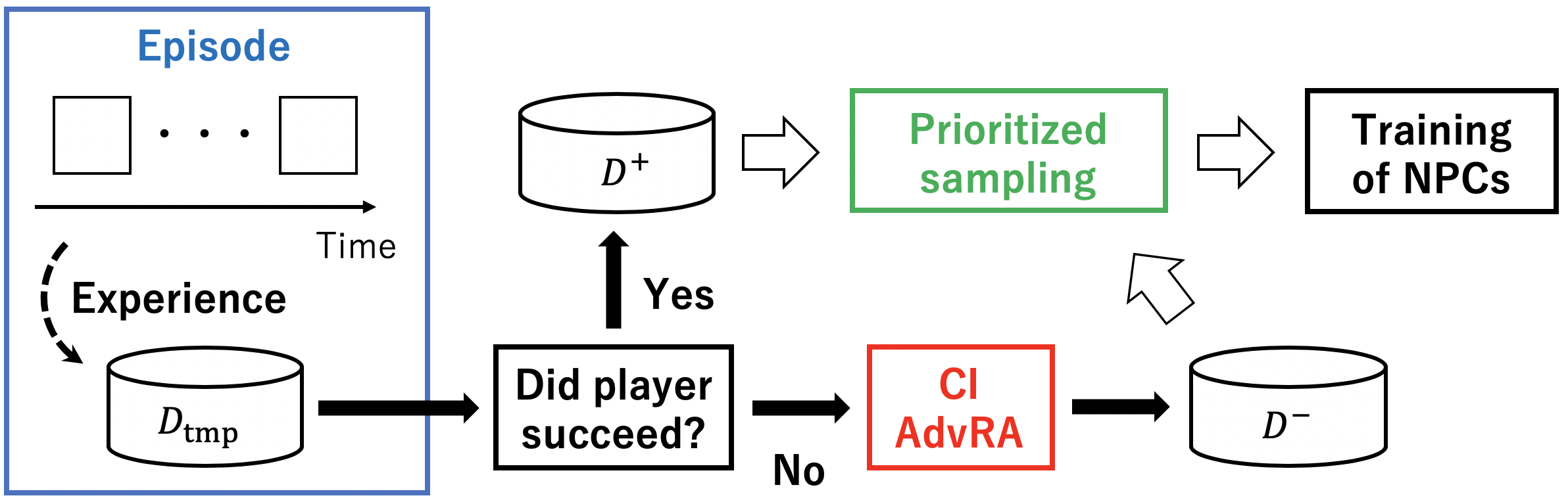}
    \caption{Conceptual image of PS-RBP.} 
    \label{fig:replay_buffer}
\end{figure}

In the case that the player rarely fails, most of the experiences in the replay buffer are ones in which the player succeeds. As a result, we are required to train adversarial NPCs with imbalanced experiences. To address this problem, we propose PS-RBP. We partition the replay buffer into two parts to separate the experience according to whether or not the player succeeds. During an episode, the experience is temporally stored in $\mathcal{D}_\text{tmp}$. After finishing the episode, the experiences in $\mathcal{D}_\text{tmp}$ are transferred to $\mathcal{D}^\text{+}$ if the player succeeds and to $\mathcal{D}^-$ if the player fails. A conceptual image is shown in Figure~\ref{fig:replay_buffer}. 

Let $e$ denote the episode number. In training the NPCs, we employ $\eta(e) \cdot S$ samples from $\mathcal{D}^\text{+}$ and $(1-\eta(e)) \cdot S$ samples from $\mathcal{D}^-$, where $\eta(e) \in [0, 1]$ is the coefficient between the number of samples from $\mathcal{D}^\text{+}$ and $\mathcal{D}^-$. $S$ is the number of samples employed in the training of the neural network.

\subsection{Overview of \algo}

\begin{figure}
    \centering
    \includegraphics[width=78mm]{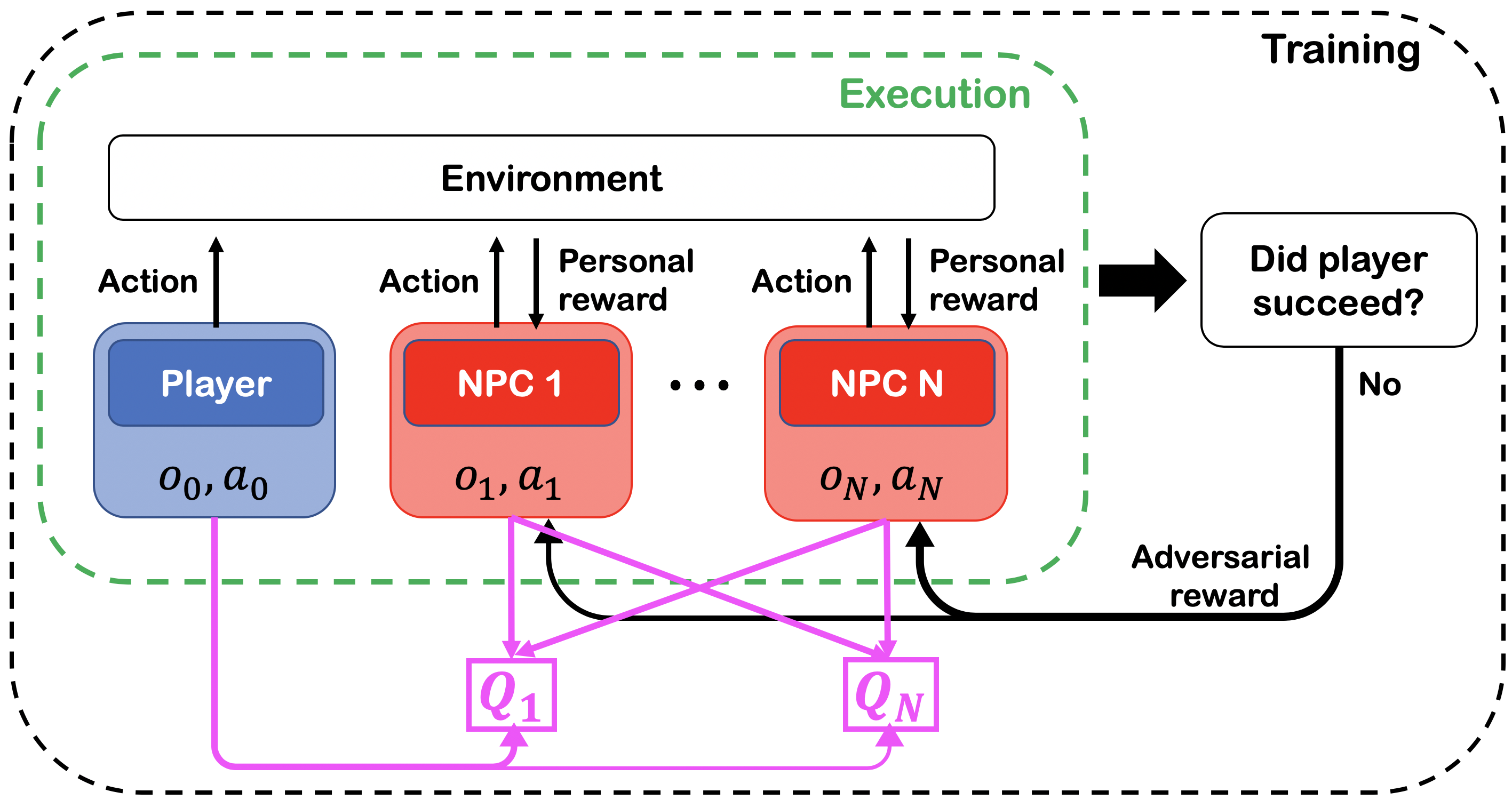}
    \caption{Problem with a player and $N$ NPCs. NPCs can get the adversarial reward when the player fails.}
    \label{fig:training_framework}
\end{figure}

An overall structure of \algo\ is shown in Figure~\ref{fig:training_framework}. As a framework for training multiple agents, we employ the MADDPG algorithm proposed in \cite{lowe2017multi}. As reviewed in Section~\ref{sec:background}, the MADDPG algorithm is a type of multi-agent PG algorithm that works well in both cooperative and competitive settings. MADDPG is a decentralized-actor-and-centralized-critic approach, which allows the critic to know the observations and policies of all agents. In this paper, we also allow the critic to know the observations and policies of the player and all NPCs. We consider applying \algo\ in the development of the player's rule; hence, this assumption is not restrictive.

Suppose that the policies of $N$ NPCs are parameterized by $\bm{\theta}=\{\theta_1, \ldots, \theta_N\}$. Also, let $\bm{\mu} = \{\bm{\mu}_1, \ldots, \bm{\mu}_N\} = \{\bm{\mu}_{\theta_1}, \ldots, \bm{\mu}_{\theta_N}\}$ be the set of all agent policies. The gradient of the expected cumulative reward for NPC $i$, $J(\theta_i)=\mathbb{E}[R_i]$ is written as:
\begin{alignat*}{2}
\nabla_{\theta_i} & J(\theta_i) = \\
&\mathbb{E}_{\bm{x}, a \sim \mathcal{D}^\pm} [ \nabla_{\theta_i} \bm{\mu}_i(o_i) \nabla_{a_i} Q_i^{\bm{\mu}}(\bm{x}, \bm{a}_{0:N}) |_{a_0=\bm{\mu}_0, a_i=\bm{\mu}_i}],
\end{alignat*}
where $\bm{a}_{0:N} = \{a_0, \ldots, a_N \}$. $\mathcal{D}^\pm$ represents the replay buffer partitioned into $\mathcal{D}^+$ and $\mathcal{D}^-$; that is, the experience is sampled from the two replay buffers using PS-RBP. Note that the experiences in $\mathcal{D}^-$ contains the reward after executing CI and AdvRA.

Using the reward function in (\ref{eq:reward_consistency}) characterized by the personal and adversarial reward, the action-value function $Q_i^{\bm{\mu}}$ is updated for all NPC $i$ as follows:
\begin{alignat*}{2}
\mathcal{L}(\theta_i) &= \mathbb{E}_{\bm{x}, a, r, \bm{x}'} [(Q_i^{\bm{\mu}}(\bm{x}, \bm{a}_{0:N})-y)^2], \\
y &= r_i + \gamma Q_i^{\bm{\mu}'}(\bm{x}', \bm{a}'_{0:N})|_{a'_0=\bm{\mu}'_0(o_0), a'_l=\bm{\mu}'_l(o_l)},
\end{alignat*}
where $\bm{\mu}' = \{\bm{\mu}_0, \bm{\mu}_{\theta'_1}, \ldots, \bm{\mu}_{\theta'_N} \}$ is the set of target policies with delayed parameters $\theta'_i$. Practically, we employ a soft update as in $\theta'_i \leftarrow \tau \theta_i + (1-\tau) \theta'_i$, where $\tau$ is the update rate.

\section{Experiments}

We present empirical results from two experiments. The first is in a multi-agent particle setting, and the second is in an autonomous driving setting.

\subsection{Simple Multi-agent Particle Environment}
\label{subsec:simple_exp}

\begin{table*}[htbp]
  \centering
  \begin{tabular}{*{12}{lrrrrrrrrrrr}}
    \toprule
    & \multicolumn{6}{c}{Simple environment} & \multicolumn{5}{c}{Autonomous driving} \\
    \cmidrule(lr){2-7}
    \cmidrule(lr){8-12}
     & \multicolumn{2}{c}{$N=1$} & \multicolumn{4}{c}{$N=3$} & \multicolumn{2}{c}{$N=1$} & \multicolumn{3}{c}{$N=2$} \\
     \cmidrule(lr){2-3}
     \cmidrule(lr){4-7}
     \cmidrule(lr){8-9}
     \cmidrule(lr){10-12}
     & Player & NPC & Player & NPC 1 & NPC 2 & NPC 3 & Player & NPC  & Player & NPC 1 & NPC 2 \\
    \midrule
    Good Agent & 2.4 & 1.3 & 5.2 & 1.6 & 1.1 & 1.3 & 3.4 & 2.9 & 5.2 & 3.0 & 2.0 \\
    Attacker & 98.9 & 98.0 & 100.0 & 98.0 & 99.5 & 99.2 & 98.0 & 98.0 & 100.0 & 98.1 & 98.2 \\
    P-Adv & 8.6 & 1.2 & 12.9 & 1.5 & 1.6 & 1.1 & 5.4 & 1.9 & 6.9 & 2.5 & 2.9 \\
    P-Adv\_AdvRA & 60.8 & 1.2 & 74.3 & 1.7 & 1.6 & 1.6 & 42.3 & 1.8 & 56.7 & 1.7 & 2.1 \\
    P-Adv\_AdvRA\_CI & 79.6 & 1.0 & 82.9 & 1.5 & 1.4 & 1.9 & 65.6 & 2.0 & 69.2 & 3.1 & 3.3 \\
    P-Adv\_PS-RBP & 11.4 & 0.9 & 23.0 & 1.4 & 1.5 & 0.9 & 8.1 & 2.6 & 8.5 & 3.6 & 3.1 \\
    \textbf{\algo} & \textbf{95.6} & \textbf{1.0} & \textbf{99.2} & \textbf{1.4} & \textbf{1.2} & \textbf{1.1} & \textbf{78.1} & \textbf{2.8} & \textbf{84.5} & \textbf{3.2} & \textbf{3.4} \\
    \bottomrule
  \end{tabular}
  \caption{Percentage of failures of player and NPC(s). \algo\ outperforms other baselines in terms of its ability to foil the player while constraining the percentage of failures of NPCs.}
\label{tab:result}
\end{table*}

We first tested \algo\ using simple multi-agent environments \cite{mordatch2017emergence}. In this simulation, we consider a player and $N$ NPCs with $N=1$ and $3$. Put simply, the player attempts to approach the goal on the shortest path. The player's goal is set as $(0, 0.7)$. However, when the player gets close to any NPCs, the player acts to maintain its distance from the NPC. 
In contrast, NPCs get the adversarial reward at the end of the episode if the player collides with another agent or a wall (i.e. a boundary) in at least one time step. 
At the same time, NPCs receive the personal reward by staying close to the goals. NPCs also try to avoid losing their personal reward; we do this to maintain the NPCs' natural behavior. More specifically, the personal reward decreases when arriving at the goal late or colliding with another agent or wall.  For the $N=1$ case, the goal of the NPC is set as $(0.5, 0)$. For the $N=3$ case, the goals are set as $(\pm 0.5, 0)$ and $(0, -0.5)$.

We implemented \algo. Our policies are parameterized by a two-layer ReLU multi-layer perceptron with 64 units per layer. In our experiment, we used the Adam optimizer with a learning rate of $0.01$. The sizes of $\mathcal{D}^+$ and $\mathcal{D}^-$ are both $1 \times 10^6$, and the batch size is set as 1024 episodes. In addition, the discount factor, $\gamma$ is set to $0.95$. For $N=1$ and $3$, $K$ is set to $K=1$ and $2$, respectively. The contribution function $g$ is defined as in (\ref{eq:contribution_func}) with $\beta = 2.0$. The weight $w$ in (\ref{eq:advra_2}) is set as $w(k) = 0.9^{k-1}$. In our simulation, through trial and error, $\eta(e)$  in PS-RBP is set as $\eta=0.5$ for $e<2500$ and $\eta=0.25$ otherwise. 
 
\paragraph{Baselines.}
We compared \algo\ with the following six baselines.

\begin{itemize}
\item \textbf{Good Agent}: Good NPCs that maximize their cumulative personal reward (i.e. only $\hat{r}$ is considered).
\item \textbf{Attacker}: Pure adversaries that maximize the cumulative adversarial reward (i.e. only $\bar{r}$ is considered).
\item \textbf{P-Adv}: Adversaries with the personal reward. 
\item \textbf{P-Adv\_AdvRA}: Adversaries with the personal reward using AdvRA (without PS-RBP and CI).
\item \textbf{P-Adv\_AdvRA\_CI}: Adversaries with the personal reward using AdvRA and CI (without PS-RBP).
\item \textbf{P-Adv\_PS-RBP}: Adversaries with the personal reward using PS-RBP (without AdvRA and CI).
\end{itemize}
We used the same parameters as in \algo\ for the above baselines. For the baselines without PS-RBP, the size of the replay buffer was $1 \times 10^6$.

\paragraph{Metrics.}
We used the following as the metrics: 1) the number of the player's failures, and 2) the number of NPCs' failures. The first metric measures the quality of being adversarial, and the second measures the naturalness (i.e. usefulness) of the adversarial situations. 

\paragraph{Results.}

\begin{figure}[t]
    \centering
    \includegraphics[width=83mm]{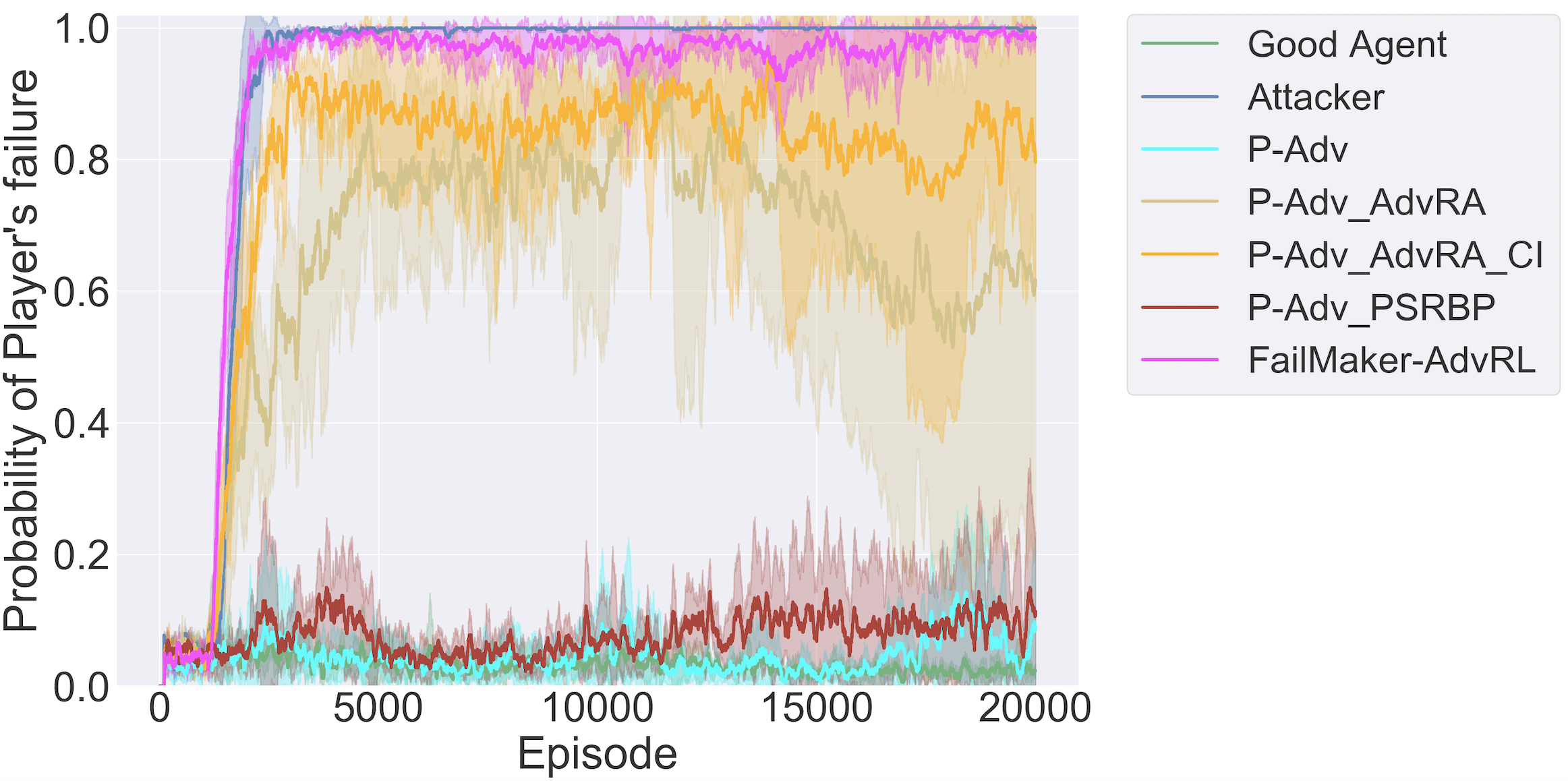}
    \caption{Comparing performance in simple multi-agent particle environment with one NPC. The player's failure rate is measured over the previous 100 episodes.}
    \label{fig:learning_curve_n1}
\end{figure}

The left half in Table~\ref{tab:result} compares the test performance of \algo\ and that of the baselines. Each value indicates the percentage of failure in 1000 episodes. For both $N=1$ and $3$, \algo\ outperforms the other baselines except for Attacker in terms of its ability to make the player fail. Attacker has a larger percentage for the player's failure, but this is because Attacker intentionally tries to collide with the player. Figure~\ref{fig:learning_curve_n1} represents the percentage of the player's failure over the number of episodes. Observe that \algo\ and Adversary achieve stable and efficient learning. Also note that P-Adv\_AdvRA and P-Adv\_AdvRA\_CI execute relatively late convergence compared with \algo.

\subsection{Autonomous Driving}
\label{sec:driving}

\begin{figure}
\centering
	\subfigure{%
		\includegraphics[clip, width=0.46\columnwidth]{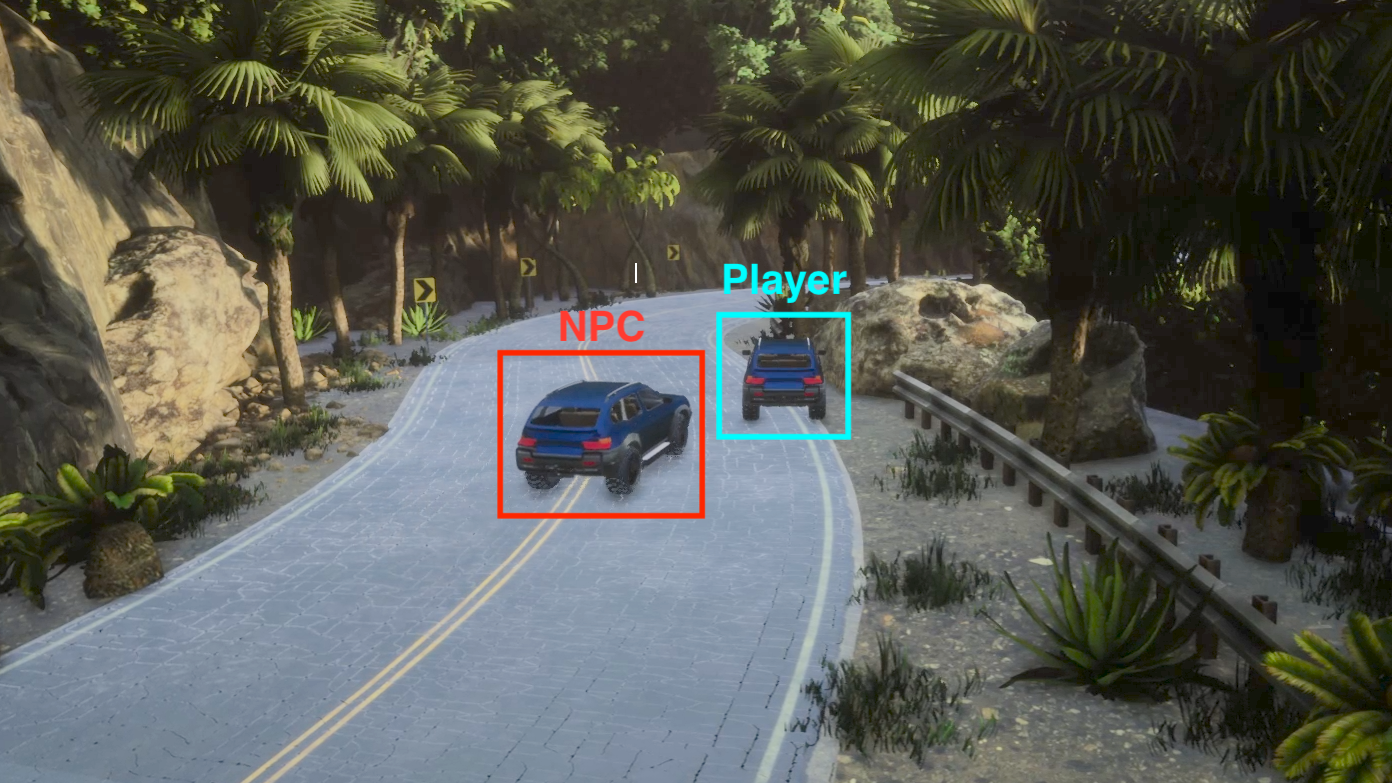}}%
	\hspace{2mm}
	\subfigure{%
		\includegraphics[clip, width=0.46\columnwidth]{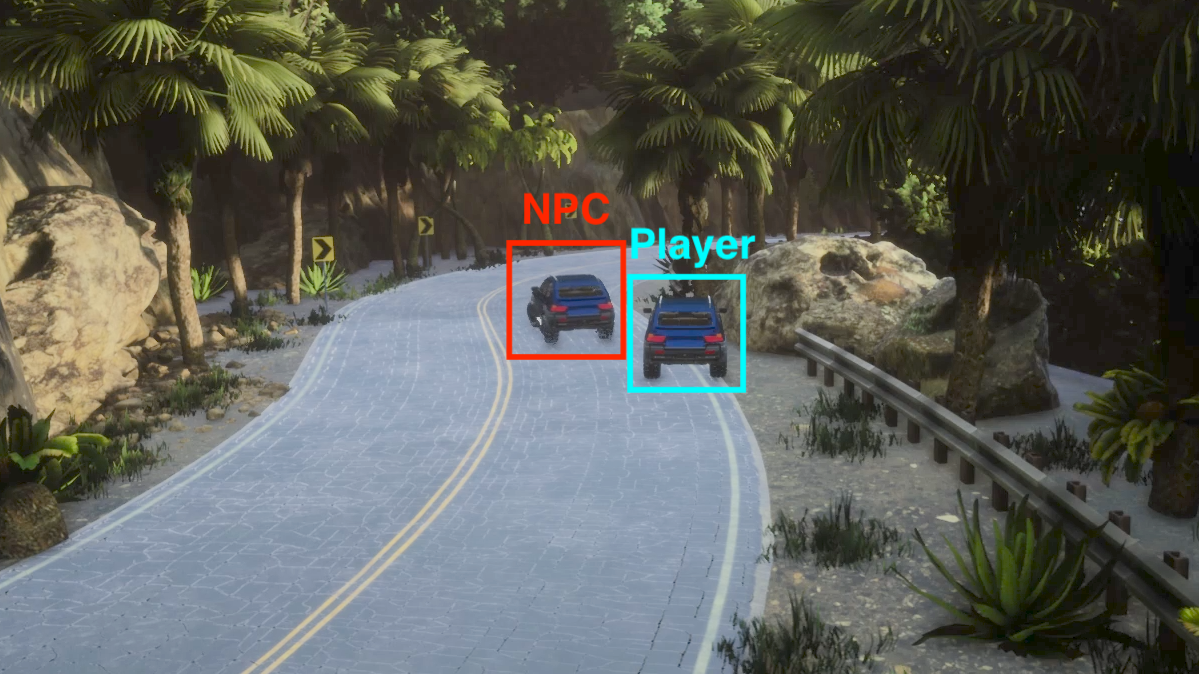}}%
	\caption{Example of failure of player (blue) induced by adversarial NPC (red).}
	\label{fig:airsim_simulation}
\end{figure}

We then applied \algo\ to the multi-agent autonomous driving problem using Microsoft AirSim \cite{airsim2017fsr}. We consider a rule-based player and $N$ adversarial RL-based NPCs. We tested for $N=1$ and $2$. In this simulation, the objective of the NPCs is to make the player cause an accident or arrive at the goal destination very late.

In this simulation, we assumed that the player and NPCs have reference trajectories, which are denoted as $\bm{\Xi}_i = \{ \bm{\xi}_i^1, \ldots, \bm{\xi}_i^T \}$, where $\bm{\xi}_i^t$ is the reference position and orientation (i.e. yaw angle) at time $t$ for agent $i$. In our experiment, the reference trajectory is obtained through manual human demonstration. We used the coastline environment in the AirSim simulator, so all the agents drive on a two-lane, one-way road. For simplicity, we assumed that 1) the reference trajectories of all agents are identical (i.e., $\bm{\Xi}_0 = \ldots = \bm{\Xi}_N$) and that 2) all agents can observe the true state at every time step. 

The player aims to follow the reference trajectory, $\bm{\Xi}_0$ using PD feedback control. Depending on the difference between the actual state $\bm{\zeta}_0^t$ and the reference state $\bm{\xi}_0^t$, the player chooses its own action. Therefore, the player follows the reference trajectory by feed-backing $\bm{\zeta}_0^t - \bm{\xi}_0^t$. The player also avoids colliding with NPCs. When the player gets closer to NPCs than a threshold, the player acts to keep its distance from the closest NPC. NPCs act to optimize their personal reward (i.e., $\hat{r}$) and adversarial reward (i.e., $\bar{r}$). The NPCs' personal reward is defined using 1) the distance with the reference trajectory and 2) the velocity. In other words, NPCs are trained to keep to the center of the read and arrive at the destination in a short time. Also, NPCs are trained to behave in an adversarial way on the basis of the adversarial reward. 

The \algo\ algorithm is implemented as follows. Our policies are parameterized by a three-layer ReLU multi-layer perceptron  with 256 units per layer. We used the Adam optimizer with a learning rate of $0.01$. The sizes of $\mathcal{D}^+$ and $\mathcal{D}^-$ are both $1 \times 10^6$, and the batch size is set as 1024 episodes. Also, for the discount factor, $\gamma$ is set to be $0.95$. 
We used the same baselines and metrics as in Section \ref{subsec:simple_exp} to evaluate \algo.

\paragraph{Results.}

The right half in Table~\ref{tab:result} compares \algo\ and the baselines. Each value indicates the failure rate of the player and NPCs in 1000 episodes. \algo\ outperforms the other baselines except for Attacker in terms of its ability to make the player fail. Figure~\ref{fig:airsim_simulation} shows an example of the simulation results using the 3D autonomous driving simulator. An adversarial NPC successfully induces the player's collision with a rock. Note that the NPC does not collide with the player or any obstacles. A sample video is in the supplemental material.

\section{Conclusion and Future Work}
We introduced a MARL problem with a rule-based player and RL-based NPCs. We then presented the \algo\ algorithm, which trains NPCs in an adversarial but natural manner. By using the techniques of CI, AdvRA, and PS-RBP, \algo\ efficiently trains adversarial NPCs.
We demonstrated the effectiveness of \algo\ through two types of experiments including one using a 3D autonomous driving simulator.

For future work, it will be important to apply our method to more realistic environments that include pedestrians or traffic signs. The first step to accomplishing this would be to craft an adversarial environment (e.g., the shape of the intersection). Also, in this work, we do not consider computer vision aspects. It will be significant to create an integrated adversarial situation while incorporating perception capabilities.

\section*{Acknowledgements}
We deeply appreciate the anonymous reviewers for helpful and constructive comments.

\bibliographystyle{named}
\bibliography{ijcai19}

\onecolumn

\section*{\Large Appendix}
\bigskip

\section*{A. \algo\ algorithm}

We provide the complete \algo\ algorithm below. 

\begin{algorithm*}[h]
\caption{\textbf{\space \algo} (Detailed version)}
\label{appdx_algorithm1}
\begin{algorithmic}[1]
\FOR {$e = 1$ to $M$}
\STATE // Execution
\STATE Initialize a random process $\mathcal{N}$ for action exploration, and then obtain the initial state $\bm{x}$
\FOR {$t = 1$ to $T$}
\STATE The player executes the action $a_0$ on the basis of the rule (i.e., $\bm{\mu}_0$).
\STATE For the adversarial NPC $i$, select the action $a_i = \bm{\mu}_{\theta_i}(o_i) + \mathcal{N}_t$
\STATE NPCs execute the actions $a=(a_1, \ldots, a_N)$ and then observe their personal reward $\hat{r}$ and new state $\bm{x}'$
\STATE Store $(\bm{x}, a, \hat{r}, \bm{x}')$ in temporal replay buffer $\mathcal{D}_\text{tmp}$
\STATE $\bm{x} \leftarrow \bm{x}'$
\ENDFOR
\smallskip \smallskip
\STATE // Reward is assigned to each NPC using CI and AdvRA depending on whether or not the player failed
\IF {the player succeeds}
\STATE Samples in $\mathcal{D}_\text{tmp}$ are moved to $\mathcal{D}^\text{+}$
\STATE $r_i^t \leftarrow \hat{r}_i^t$
\ELSE
\STATE Samples in $\mathcal{D}_\text{tmp}$ are moved to $\mathcal{D}^-$
\STATE Identify the contributors with CI and obtain $\mathcal{C}_k (k=1,2,\ldots,K)$
\STATE Allocate the adversarial reward to each contributor with AdvRA
\STATE $r_i^t \leftarrow \hat{r}_i^t + \alpha \cdot \bar{r}_i^t$ 
\ENDIF
\smallskip \smallskip
\STATE // Training of NPCs
\FOR {NPC $i=1$ to $N$}
\STATE Randomly choose $\eta(e) S$ samples from $\mathcal{D}^\text{+}$ and $(1-\eta(e))S$ samples from $\mathcal{D}^-$, and create a random minibatch of $S$ samples $(\bm{x}^j, a^j, r^j, \bm{x}'^j)$ using PS-RBP
\STATE Set $y^j = r_i^j + \gamma Q_i^{\bm{\mu}'}(\bm{x}', a_0', a_1', \ldots, a'_N)|_{a'_0=\bm{\mu}'_0(o_0^j), a'_l=\bm{\mu}'_l(o_l^j)}$
\STATE Update critic by minimizing the loss:
\[
\mathcal{L}(\theta_i) = \frac{1}{S}\sum_j \left(y^j - Q_i^{\bm{\mu}}(\bm{x}^j, a_0^j, a_1^j, \ldots, a^j_N) \right)^2
\]
\STATE Update actor using the sampled policy gradient:
\[
\nabla_{\theta_i} J \approx \frac{1}{S} \sum_j \nabla_{\theta_i} \bm{\mu}_i(o_i^j) \nabla_{a_i} Q_i^{\bm{\mu}}(\bm{x}^j, a_0^j, a_1^j, \ldots, a_N^j) |_{a_0=\bm{\mu}_0(o_0^j), a_i=\bm{\mu}_i(o_i^j)}
\]
\ENDFOR
\STATE Update target network parameters for each NPC $i$:
\[
\theta_i' \leftarrow \tau \theta_i + (1-\tau)\theta_i'
\]
\ENDFOR
\end{algorithmic}
\end{algorithm*}

\end{document}